\documentclass[10pt,twocolumn,letterpaper]{article}

\usepackage{cvpr}
\usepackage{times}
\usepackage{epsfig}
\usepackage{graphicx}
\usepackage{amsmath}
\usepackage{amssymb}
\usepackage{booktabs}
\usepackage{mwe}
\usepackage{xspace}

\usepackage[dvipsnames]{xcolor}
\usepackage{subcaption}

\usepackage[pagebackref=true,breaklinks=true,letterpaper=true,colorlinks,bookmarks=false]{hyperref}

\cvprfinalcopy

\ifcvprfinal\fi
\begin{document}

\title{SPLAT: Semantic Pixel-Level Adaptation Transforms for Detection }

\author{Eric Tzeng\\
UC Berkeley\\
{\tt\small etzeng@eecs.berkeley.edu}
\and
Kaylee Burns\\
UC Berkeley\\
{\tt\small kayleeburns@berkeley.edu}
\and
Kate Saenko\\
Boston University\\
{\tt\small saenko@bu.edu}
\and
Trevor Darrell\\
UC Berkeley\\
{\tt\small trevor@eecs.berkeley.edu}
}

\maketitle
\newcommand{\lossGAN}{\mathcal{L}_{\text{GAN}}\xspace}
\newcommand{\lossCycle}{\mathcal{L}_{\text{cycle}}\xspace}
\newcommand{\lossTask}{\mathcal{L}_{\text{task}}\xspace}
\newcommand{\lossSA}{\mathcal{L}_{\text{SA}}\xspace}
\newcommand{\lossPreserveLabel}{\mathcal{L}_{\text{label}}\xspace}
\newcommand{\lossSPLAT}{\mathcal{L}_{\text{SPLAT}}\xspace}
\newcommand{\StoT}{{s \rightarrow t}\xspace}
\newcommand{\TtoS}{{t \rightarrow s}\xspace}

\begin{abstract}
Domain adaptation of visual detectors is a critical challenge, yet existing methods have overlooked  pixel appearance transformations, focusing instead on bootstrapping and/or 
domain confusion losses.
We propose a Semantic Pixel-Level Adaptation Transform (SPLAT) approach to detector adaptation that efficiently generates cross-domain image pairs. Our model uses aligned-pair and/or pseudo-label losses to adapt an object detector to the target domain, and can learn transformations with or without densely labeled data in the source (e.g. semantic segmentation annotations).  Without dense labels, as is the case when only detection labels are available in the source, transformations are learned using CycleGAN alignment.  Otherwise, when dense labels are available we introduce a  more efficient cycle-free method, which exploits pixel-level semantic labels to condition the training of the transformation network. The end task is then trained using detection box labels from the source, potentially including labels inferred on unlabeled source data. We show both that pixel-level transforms outperform prior approaches to detector domain adaptation, and that our cycle-free method outperforms prior models for unconstrained cycle-based learning of generic transformations while running 3.8 times faster. Our combined model improves on prior detection baselines by 12.5 mAP adapting from Sim 10K to Cityscapes, recovering over 50\% of the missing performance between the unadapted baseline and the labeled-target upper bound.
\end{abstract}

\section{Introduction}

\begin{figure}
\centering
\includegraphics[width=\columnwidth]{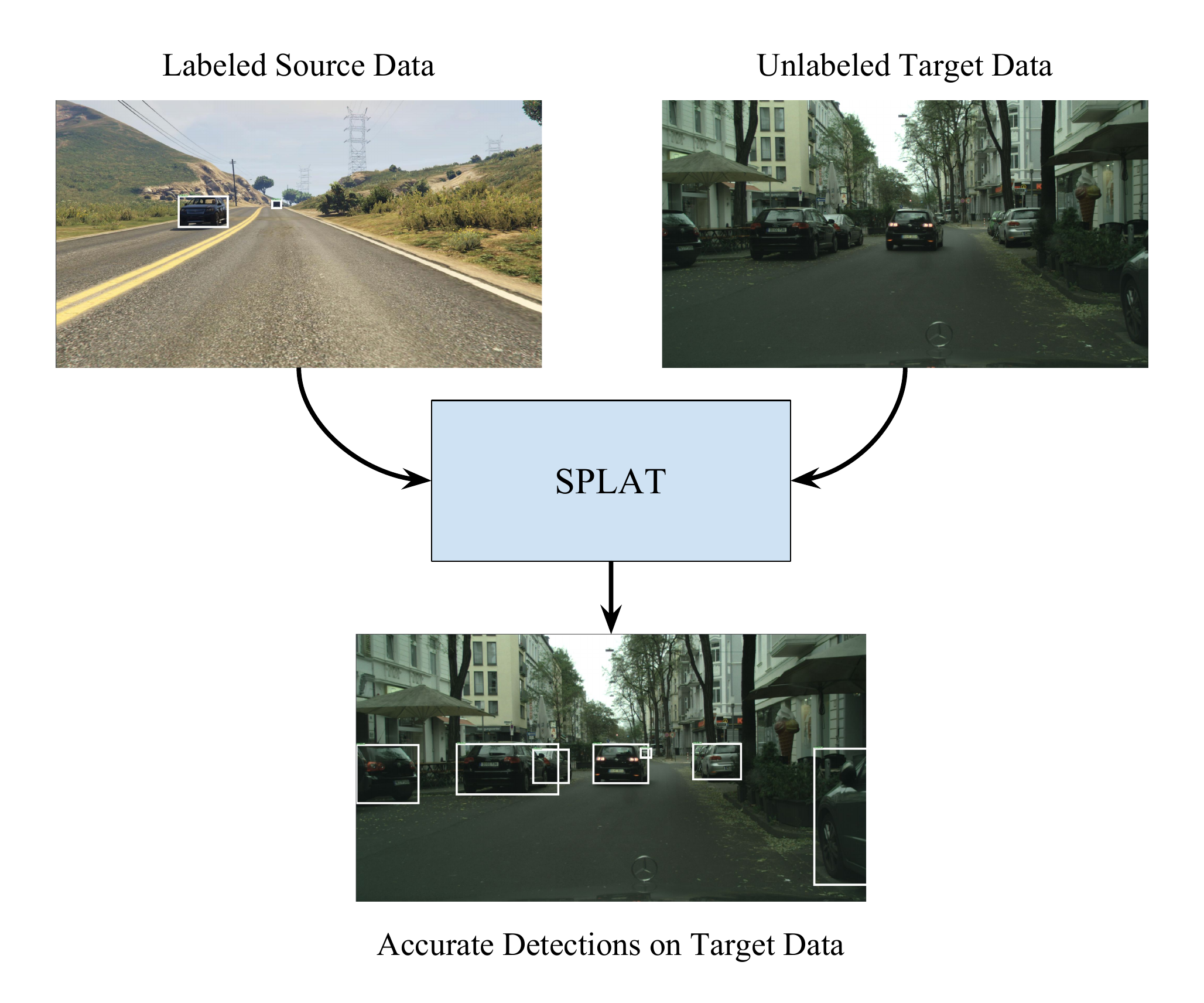}
\caption{We propose a new method, SPLAT, that tackles the task of unsupervised domain adaptation for detection. SPLAT is trained on labeled source data and unlabeled target data, and learns to produce accurate detections in the target domain. We improve detection results by 12.5 points over previous methods, setting the new state of the art. }
\label{fig:teaser}
\end{figure}

To realize the promise of computer vision in practice, machine learning models must learn to adapt to shifting data distributions quickly and gracefully.
The specific goal of adapting driving scene object detectors from synthetic  
to real scenes 
has become important as many groups are attempting to learn driving policies in a simulated environment and transfer them to a real world environment.

Image and pixel-level adaptation of image classifiers are relatively well studied (see citations below), 
but detection adaptation methods have recieved less attention. Existing detection adaptation approaches either pre-date deep learning~(e.g.,~\cite{Donahue_2013_CVPR})
or employ a domain classification loss on the proposal generating layer and box classifiction network head to align feature distributions across domains \cite{chen2018domain}.    
\begin{figure*}
    \centering
    \includegraphics[width=\textwidth]{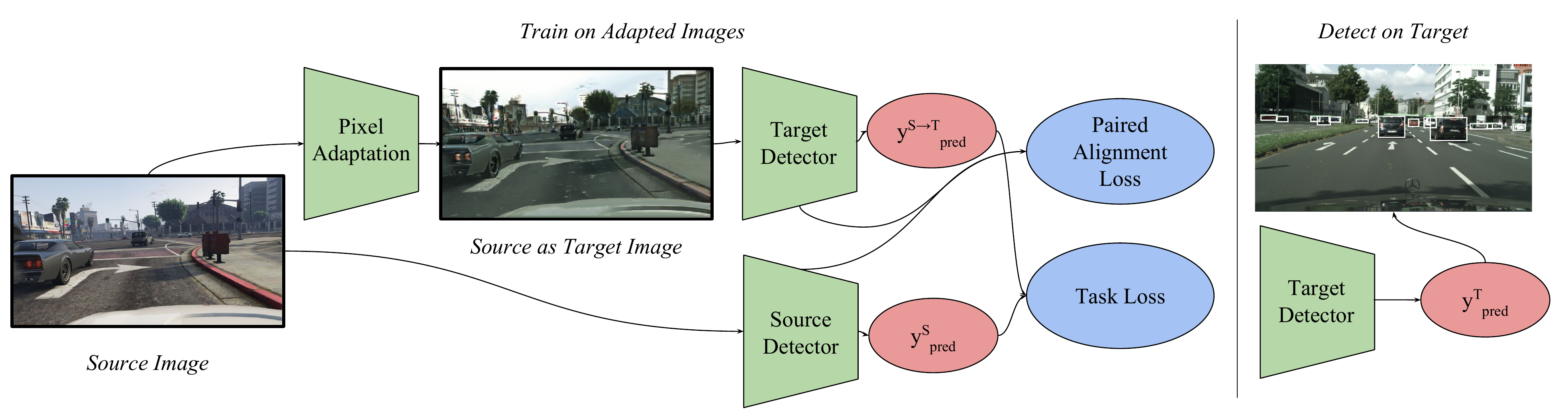}
    \caption{An overview of SPLAT, our proposed domain adaption technique for object detection. Our method leverages
    fast pixel adaptation when 
    semantic segmentation labels are available in the source, and can also use slower, cycle-based adaptation techniques when they are not. We pair the adapted images with their corresponding source version to train a target detector by aligning the features of the source and target models, and/or training a task loss with labels inferred from source data. On the far right, we show some detection results on the target dataset, Cityscapes.}
\label{fig:dda}
\end{figure*}

When evaluated on benchmark challenges for driving detection domain adaptation, existing methods perform poorly, e.g., recovering less than half of the domain shift when adapting from \textit{Sim10k}~\cite{johnson_roberson2017} (a synthetic detection dataset generated from Grand Theft Auto V) to \textit{Cityscapes}~\cite{cordts2016_cityscapes}. 
One potential reason is that feature distribution alignment is suboptimal when samples include many background patches or whole-image scenes that differ in object layout between domains.
In contrast, adaptation approaches based on learning pixel-based image-to-image transforms \cite{Hoffman_cycada2017} have shown promise for semantic segmentation tasks, recovering up to 84\% of source-to-target loss, but this class of methods have not been heretofore reported for the task of detection.  

We propose an approach to detection domain adaptation based on pixel-transformed source-to-target imagery. Our Semantic Pixel-Level Adaptation Transform (SPLAT) method efficiently generates cross-domain image pairs and employs pseudo-label losses and/or alignment losses on paired images when training a target domain detector. Figure \ref{fig:teaser} summarizes the problem statement and Figure~\ref{fig:dda} overviews our general approach. Our experiments confirm that pixel-level transformations provide a significant boost for detection adaptation compared to previous methods using domain-confusion losses and outperform all published baselines on the benchmark \textit{Sim10k-to-CityScapes} challenge by a large margin.

State-of-the-art pixel transformations for domain adaptation have often been based on cycle-transforms, e.g., as used in Cycle-GAN~\cite{zhu_arxiv17} and CyCADA~\cite{Hoffman_cycada2017}. 
However cycle-consistency comes at a price; Cycle-GAN based optimization has two separate transformers that must be learned simultaneously, and optimizing them requires four image generations per iteration. Yet fully bi-directional reconstruction is not necessary for detection adaptation, i.e.,  target-to-source transformations are not used at inference time in our final method.  

We therefore propose a novel cycle-free approach to learning the pixel transform in our method. A transform cannot be learned reliably from unlabeled data without cycles, so we  leverage available additional side labels on the source domain, specifically semantic-segmentation labels, to learn the transformation. Such annotations are easy to generate for synthetic data. We add a novel semantic pixel prediction loss to form our \textit{SPLAT-lite} model, as described below, which provides constraints sufficient to efficiently learn effective pixel transforms for detection adaptation. 

Below we show results of our general \textit{SPLAT} approach and our 
fast \textit{SPLAT-lite} method on a benchmark challenge for object detection adaptation problems, adapting from Sim 10K to Cityscapes.  Our results confirm both the superiority of a pixel-transform approach for detection adaptation, and the greater efficiency of our cycle-free \textit{SPLAT-lite} model for learning transforms for adaptation.

\section{Related work}
\label{sec:related}

Domain adaptation was first investigated as a task for visual recognition by~\cite{saenko_eccv10} to better understand and address the dramatic loss of performance on out-of-domain data. Visual differences between domains were shown to lead to degraded performance when models trained in one domain were na\"ively applied to other domains due to so-called domain shift or dataset bias~\cite{efros_cvpr11}.

Early work focused on closing this gap for classification tasks at the object- or pixel-level (i.e. the tasks of classification and semantic segmentation), where each class is well represented in the set of labels. The success of these techniques have relied on two ideas, used in isolation or in concert: enforcing similarity of features deep within  convolutional networks or learning pixel-level transformations from the source to the target that can be used as additional data during training.

Alignment approaches incorporate measures of distance between source and target feature distributions as a part of the task loss. The first of these methods looked at the differences between mean or variance as distribution distance metrics~\cite{long_icml15,sun_taskcv16}. Other more recent approaches approximate distribution differences by learning a domain discriminator that classifies which features are from the source and which features are from the target~\cite{ganin_icml15,tzeng_iccv15,tzeng_cvpr17}.
Once the distance between domains can be measured in this way, a deep representation can be adversarially learned to directly minimize the effects of domain shift.

We can draw parallels between these adversarial adaptation approaches and generative adversarial networks (GANs)~\cite{goodfellow_nips14}; much like how adversarial adaptation techniques learn a representation that fools an adversarial domain discriminator, GANs optimize a generator to output images that fool an adversarial discriminator attempting to distinguish real from generated images.
A variety of effort has been devoted to various techniques for stabilizing GANs, including but not limited modifications to the training protocol~\cite{salimans2016improved}, alternatives to the minimax loss function~\cite{arjovsky_arxiv17,gulrajani_nips17,mao_iccv17}, and normalization of network parameters~\cite{miyato_iclr18}, to name a few.
As they improved, GANs also began seeing widespread success in a variety of applications, including image generation~\cite{denton2015deep,radford2015unsupervised,zhao2016energy}, image editing~\cite{zhu2016generative}, and even unsupervised or semi-supervised representation learning for standard computer vision tasks~\cite{salimans2016improved,donahue2016adversarial,dumoulin2016adversarially}.

Another class of adaptation methods attempt to perform adaptation directly using the pixels of an image, rather than relying exclusively on intermediate feature spaces.
Deep Reconstruction-Classification Networks~\cite{ghifary_eccv16} adapt by learning a shared encoding that is trained to label source imagery but directly reconstruct target images.
CoGANs~\cite{cogan} learn a partially-shared pair of GANs that output aligned source/target pairs of images when provided with a single output, and apply this model to domain adaptation.

The last group of adaptation approaches we discuss are methods that learn a transformation to redraw source images in a style that matches the target distribution.
\cite{shrivastava_cvpr17} learns a GAN-based model to refine synthetic images into more realistic training data such that the L1 distance to the original synthetic image is minimized.
\cite{bousmalis_cvpr17} also attempts to redraw source images into target images while learning a task network that can successfully label both source and source-as-target images.
Finally, CyCADA~\cite{Hoffman_cycada2017} extends CycleGAN to domain adaptation by augmenting it with label information and performing an additional round of feature adaptation to transform source images while maintaining their semantic content.
This class of approaches is most similar to our method, and so we recap relevant details here.

\section{Background}
\label{sec:background}

Pixel adaptation approaches model the target distribution from the source dataset to take advantage of source labels. Many recent methods apply adversarial adaptation losses directly to the pixels of images, effectively redrawing source images in the style of the target domain. The most basic way of performing this pixel-level adaptation is to simply train a GAN such that the generator $G$ takes source images $X_s$ as input and attempts to match its output to the distribution of target images $X_t$:

\begin{equation}
  \begin{split}
  \lossGAN(G, D, X_s, X_t) = &\mathbb{E}_{x_t \sim X_t}[\log D(x_t)] \\
  + &\mathbb{E}_{x_s \sim X_s}[\log(1 -  D(G(x_s)))].
  \end{split}
  \label{eq:ganLoss}
\end{equation}

Previous methods achieving strong adaptation performance relied on very computationally expensive content preservation losses.
CycleGAN~\cite{zhu_arxiv17} uses a content preservation loss relying on a reverse-direction generator $G'$ and discriminator $D'$ that adapt target samples into source samples, trained adversarially in a similar manner to $G$ and $D$.
The generators are constrained to be inverses of each other via a cycle-consistency loss:

\begin{equation}
\begin{split}
  \lossCycle(G, G', X_s, X_t) = \mathbb{E}_{x_s \sim X_s} &||G'(G(x_s)) - x_s||_1 \\
  + \mathbb{E}_{x_t \sim X_t} &||G(G'(x_t)) - x_t||_1.
\end{split}
\label{eq:cycleLoss}
\end{equation}

Because CycleGAN was not originally proposed as a domain adaptation method, it was not designed to make use of the task labels $Y_s$.
The authors of CyCADA \cite{Hoffman_cycada2017} extend CycleGAN to an adaptation setting by proposing an additional semantic alignment loss.
This second component introduces a source task network $f_s$ that is trained on the labeled source data $(X_s, Y_s)$.
This task network is used to regularize the generator $G$ so as to minimize the task loss on the adapted data $\mathcal{L}_T(T(G(x_s)), x_y)$.

\begin{equation}
\begin{split}
  \lossSA(G, G', f_s, X_s, X_t) = &\lossTask(f_s, G(X_s), f_s(X_s))\\
  + &\lossTask(f_s, G'(X_t), f_s(X_t))
\end{split}
\end{equation}
where $\lossTask(f, X, Y)$ denotes the task loss (typically cross-entropy) of the task network $f$ on samples $X$ against the labels $Y$.

By adapting source images to the target domain, we can effectively apply supervisory information from the source domain to training examples that closely resemble the target unlabeled data, thereby leading to improved target performance.

Generating images $X_\StoT = \{G(x_s) \mid x_s \in X_s\}$ that resemble the target images $X_t$ in appearance but contain the content from the source images $X_s$ allows us to form a new, pseudo-labeled dataset $(X_\StoT, Y_s)$.
When the cross-domain  generator $G$ successfully accomplishes this goal, training a task model on this pseudo-labeled dataset can lead to strong performance in the target domain.

However, previous work~\cite{Hoffman_cycada2017} has shown that the GAN objective in Equation~\ref{eq:ganLoss} alone can often train a poorly conditioned generator.
Left unconstrained, the generator produces adapted source samples $X_\StoT$ that can vary wildly in content from the original source samples.
In turn, this means that the original labels $Y_s$ no longer correspond to the adapted samples.
Thus, training a task model on $(X_\StoT, Y_s)$ leads to poor task performance.
In order to prevent deviation from the label information $Y_s$, it is imperative to constrain $G$ with an additional loss, such that $G$ is encouraged to preserve the semantic content in $X_s$ during pixel-level adaptation.

\section{Semantic Pixel-Level Adaptation Transforms}

We propose a method called Semantic Pixel-Level Adaptation Transforms (SPLAT) that performs pixel transformations from source to target to tackle the object detection adaptation problem.
SPLAT is flexible and can optionally make use of additional label information such as segmentation labels when such data is available.
In turn, the ability to make use of this extra information leads to a more efficient cycle-free pixel adaptation method that both runs faster than existing approaches and allows us to learn detection models that are more accurate in the target domain.

Existing pixel adaptation methods have demonstrated that it is possible to directly transform source images into target images in order to learn task models that are effective in the target domain.
However, their applications were limited primarily to the realm of classification and segmentation.
We show for the first time that these methods can be applied to detection domain adaptation, producing a cross-domain aligned set of images.
We demonstrate that pixel-adaptation methods can be used to learn target domain object detectors that are robust to the negative effects of domain shift.

First, we learn a source-to-target generator $G$ (e.g., via cycle constraints as in ~\cite{Hoffman_cycada2017}), and use this generator on the source images $X_s$ to produce adapted source images $X_\StoT = G(X_s)$.
This provides us a pair-aligned set of images, $X_s$ and $X_\StoT$, such that each image $x_s$ in $X_s$ has a counterpart $x_\StoT = G(x_s)$ containing the same content in a different style in $X_\StoT$.

\subsection{Pseudo-labeling and pair alignment}

We explore two methods for using these aligned datasets to learn a target task network: pseudo-labeling and explicit pair alignment.
These methods can be used either independently or together at the same time.

The simplest way of using these two image sets together is to use the adapted images $X_\StoT$ along with the original source labels $Y_s$ in order to form a surrogate, ``pseudo-labeled'' dataset that closely resembles the true target data.
We can then learn a target task network $f_t$ that minimizes
\begin{equation}
  \mathcal{L}_\text{pseudo}(f_t, X_\StoT, Y_s) = \lossTask(f_t, X_\StoT, Y_s)
\end{equation} on this newly formed dataset, where once again $\lossTask$ corresponds to the task loss---in this case, a combination of cross-entropy for box classification and a regression loss for refining predictions.

We also propose a method to optionally pseudo-label an unlabeled source dataset $X_{s'}$ when such data is available.
We utilize a source network $f_s$ that is trained on the available labeled source dataset $(X_s, Y_s)$.
Since labeled source data is plentiful, it is reasonable to expect that inferred labels $\hat{Y}_{s'}=f_s(X_{s'})$ will be fairly accurate.
Thus, even despite the lack of true ground truth data, we can still form a pseudo-labeled target dataset $(G(X_{s'}), \hat{Y}_{s'})$.
We can then use this dataset to learn a target network $f_t$.
This process as described attempts to use an unlabeled source dataset $X_{s'}$ to minimize the loss
\begin{equation}
  \mathcal{L}_{\text{unsup}}(f_s, f_t, G, X_{s'}) = \lossTask(f_t, G(X_{s'}), f_s(X_{s'})).
\end{equation}
This enables our model to use as much source data as is available, even if that source data is unlabeled or labeled for another task such as segmentation.
This process is depicted in Figure~\ref{fig:dda}.

We also explore the possibility of using corresponding pairs of images from $X_s$ and $X_\StoT$ to constrain our target task network $f_t$ with an additional /emph{pair alignment loss}:
\begin{equation}
\mathcal{L}_{\text{pair}}(\phi_s, \phi_t, G, X_s) = \mathbb{E}_{x_s \sim X_s} || \phi_s(x_s) - \phi_t(G(x_s)) ||^2_2.
\end{equation}
This loss enforces that the target network's intermediate features $\phi_t$ should closely resemble the intermediate features $\phi_s$ of the original source network $f_s$ when applied to paired images.

These two methods of learning from adapted images $X_\StoT$ are complementary, and can be used independently or in concert.
In Section~\ref{sec:results-detection} we evaluate the effect each of these loss functions has on our final model performance.

\subsection{Cycle-free pixel adaptation }

\begin{figure}
    \centering
    \includegraphics[width=0.48\textwidth]{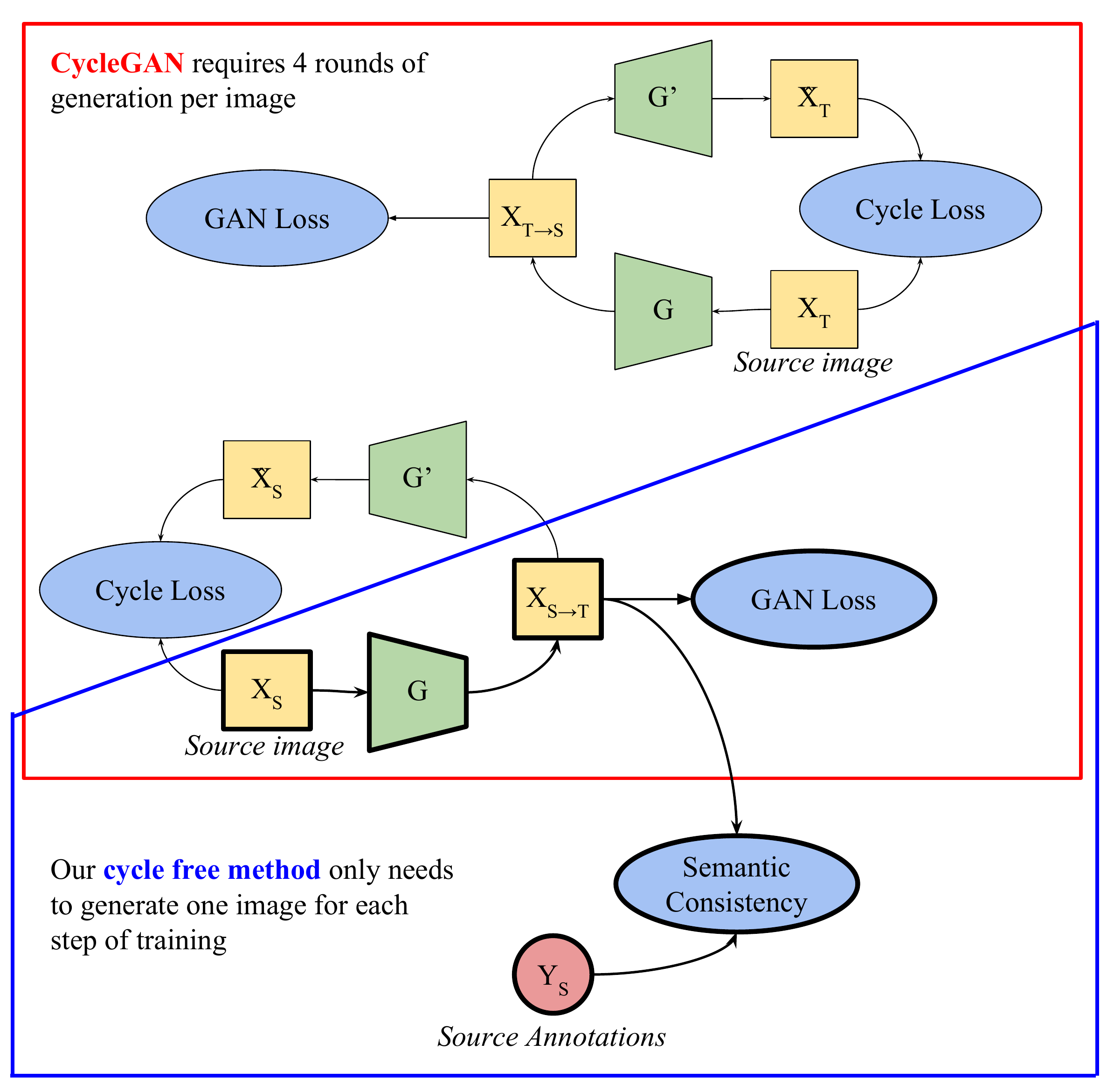}
    \caption{
    A visual comparison between our proposed cycle-free pixel adaptation method (used in \textit{SPLAT-lite}) and CycleGAN.
    By incorporating source annotations while learning our pixel transformer, we can eliminate many of the components present in previous methods.
    In particular, we note that our model requires one-fourth as many image generation passes as CycleGAN does per iteration.
    In practice, using this label information leads to greatly improved efficiency and stronger performance on the final task.}
 \label{fig:model}
\end{figure}

Our detection adaptation method is agnostic to the specific pixel transformation model that is used.
In practice, performing detection with existing pixel-level adaptation methods such as CyCADA yields surprisingly strong performance.
However, these models can be quite difficult to train and do not make use of the semantic labels present in adaptation tasks.
Thus, as an additional contribution, we introduce a novel, lightweight pixel-level adaptation method that eliminates the cumbersome cycle loss in favor of a loss that incorporates label information.

Existing pixel-adaptation methods such as CycleGAN/CyCADA produce impressive transformed images; however, training such models is often computationally quite expensive.
For example, as explained in Section~\ref{sec:background}, in order to compute the cycle loss used in CycleGAN, each iteration we must perform four image generation passes: $G(x_s)$, $G'(x_t)$, $G'(G(x_s))$, and $G(G'(x_t))$, as well as an additional discriminator update to train $G'$.
To prevent content from changing during pixel transformation, CyCADA uses a semantic consistency loss that requires two additional task network passes, $T(x_s)$ and $T(x_t)$.
Running this many networks each iteration quickly becomes prohibitively expensive.
Indeed, the authors of CyCADA~\cite{Hoffman_cycada2017} note that they were forced to drop the semantic consistency $\lossSA$ due to memory constraints.

We argue that such cumbersome models are overkill for adaptation, especially when pixel-level semantic segmentation labels are available for the source domain.
As an alternative, we propose a much more lightweight method for pixel-level adaptation.
By directly incorporating available pixel-level semantic labels $Y_s$, we can devise a novel alternative loss that effectively constrains the source-to-target generator $G$ without requiring the use of a reverse-direction generator $G'$.
In particular, we propose constraining $G$ such that the source task network $T$ is still able to predict the ground-truth labels $Y_s$ on the adapted images $X_\StoT$:
\begin{equation}
  \begin{split}
    \lossPreserveLabel(G, T, X_s, Y_s) = \lossTask(T, G(X_s), Y_s).
  \end{split}
\end{equation}

Our lightweight model eliminates the computationally expensive cycle loss, and instead combines the GAN loss for image appearance alignment with the novel label preservation loss. This enables content preservation during domain translation and leads to efficient pixel-level domain adaptation.
The resulting pixel-adapted imagery can then be used to form a new training set $(X_\StoT, Y_s)$ and train a task network via either simple pseudo-labeling or pair alignment, as described above. We call this version of our model \textit{SPLAT-lite}.
In full, our cycle-free pixel adaptation method optimizes the loss function
\begin{equation}
  \begin{split}
    \lossSPLAT(G, T, X_s, X_t, Y_s) = &\lossGAN(G, D, X_s, X_t) \\
    + &\lossPreserveLabel(G, T, X_s, Y_x).
  \end{split}
\end{equation}

Figure~\ref{fig:model} visually depicts our cycle-free method alongside a comparison to CycleGAN.
It is visually apparent how much simpler our model is compared to previous work.
In practice, this simplified training scheme leads to faster training times, since less computation is required for each training iteration.
In addition, since our proposed method only uses one adversarial discriminator, optimization is simpler and proceeds in a more stable manner.

\section{Experiments}
We focus our evaluation of our proposed method on a popular and challenging synthetic-to-real car detection challenge.
We begin with an overview of our experimental setup and architecture design in Sections~\ref{sec:results-data} and \ref{sec:results-arch}.
Next, we report our results on detection adaptation, comparing it with previous adaptation approaches and discussing how well feature- and image-level adaptation techniques generalize to detection in  Section~\ref{sec:results-detection}.
To better understand the effect of our cycle-free pixel transformer, we then ablate SPLAT with different pixel adaptation techniques, and evaluate the relative speed and accuracy of these approaches in Section~\ref{sec:results-pixel}.

\subsection{Datasets}
\label{sec:results-data}
We test detection adaptation performance from synthetic to real driving scenarios. Our source domain consists of Grand Theft Auto V imagery and our target domain is real-world driving data captured from dashboard mounted cameras. Consistent with prior work, we train our source detection model on \textit{Sim10k} \cite{johnson_roberson2017}, a synthetic detection dataset generated from Grand Theft Auto V, and test on \textit{Cityscapes}\cite{cordts2016_cityscapes}.  \textit{Sim10k} contains 10,000 images and 58,701 bounding boxes, all of which are used to train our adaptation model. There are no bounding box annotations \textit{Cityscapes}, so we construct tight bounding boxes around instance-level segmentations. We treat the 2975 training images as our target dataset and evaluate on the 500 validation images. We only test detection results for car.

Our model is amenable to training with additional source data that is either unlabeled or has labels for a different task. In our experiments, we take advantage of an additional in-domain source dataset with segmentation labels to improve the speed of our pixel-adaptation model. We use a distinct set of images generated from Grand Theft Auto V and annotated with semantic segmentation labels. This is the \textit{GTA5} dataset~\cite{richter_eccv16}. It contains 24966 images that we use to train the pixel adaptation component of SPLAT-Lite. In our pixel adaptation analysis, we describe the training details of SPLAT-lite and show that, by compromising training speed, we can achieve comparable accuracy without additional source data.

\subsection{Implementation details}
\label{sec:results-arch}
We use Faster R-CNN \cite{renNIPS15fasterrcnn} as our detection module with the hyperparameters suggested in the authors' implementation~\cite{Detectron2018}. We scale each image to a height of 512 pixels and a maximum width of 1024 pixels to fit in GPU memory. Our backbone architecture is the ResNet-based feature pyramid network~\cite{lin_cvpr17}, and we initialize from weights pretrained on ImageNet~\cite{ILSVRC15}. We report adaptation results for our model using a cycle-free pixel transformation model, and opt not to include a pair alignment loss.

\begin{table}
    \centering
    \begin{subtable}{0.48\columnwidth}
    \centering
    \scriptsize
    \begin{tabular}{c}
    \toprule\midrule
    RGB image $x \in \mathbb{R}^{256\times256\times3}$ \\ \midrule
    ResBlock down 64 \\ \midrule
    ResBlock down 32 \\ \midrule
    ResBlock up 16 \\ \midrule
    ResBlock up 8 \\ \midrule
    BN, ReLU, 3$\times$3 conv 3, Tanh \\ \midrule
    \bottomrule
    \end{tabular}
    \caption{Cycle-free generator}
    \end{subtable}
    \begin{subtable}{0.48\columnwidth}
    \centering
    \scriptsize
    \begin{tabular}{c}
    \toprule\midrule
    RGB image $x \in \mathbb{R}^{256\times256\times3}$ \\ \midrule
    ResBlock down 8 \\ \midrule
    ResBlock down 16 \\ \midrule
    ResBlock down 32 \\ \midrule
    ResBlock down 64 \\ \midrule
    ResBlock down 128 \\ \midrule
    ResBlock down 128 \\ \midrule
    ReLU \\ \midrule
    Global sum pooling \\ \midrule
    Dense $\rightarrow$ 1 \\ \midrule
    \bottomrule
    \end{tabular}
    \caption{Cycle-free discriminator}
    \end{subtable}
    \caption{The cycle-free generator and discriminator architectures used in our SPLAT-lite experiments. We model our architectures after SN-GAN, but use spectral normalization in both the generator and the discriminator. Architectures are described using the same notation as~\cite{miyato_iclr18}; see the original paper for details.}
    \label{tab:splat-arch}
\end{table}  

Our cycle-free pixel adaptation transformer in our primary results is trained on the \emph{GTA5} dataset.
The generator and discriminator are variants of architectures used by SN-GAN~\cite{miyato_iclr18}, but modified to perform image-to-image generation---the exact architectures we use are outlined in Table~\ref{tab:splat-arch}.
The task network used in the label preservation loss is an FCN-8s~\cite{long_cvpr15} fully convolutional network, again trained on the \emph{GTA5} dataset.
In Section~\ref{sec:results-pixel}, we also experiment with variants of the SPLAT architecture and compare it with previous pixel-adaptation methods to validate our claims of improved performance and efficiency.

Our training procedure is as follows. We begin by training a source Faster R-CNN model on \textit{Sim10k}. That model is used to generate detections on the \textit{GTA5} dataset. Next, we accept all proposals with a confidence of 0.5 or greater and treat them as ground truth annotations for \textit{GTA5}. We transform \textit{GTA5} into the target \textit{Cityscapes} style using our learned pixel transformer. Finally, we train Faster R-CNN using the transformed images and the detections inferred by the source detectors on the original \textit{GTA5} images.

\subsection{Detection adaptation results}
\label{sec:results-detection}

\begin{table}
    \centering
    \begin{tabular}{lc}
    \toprule
    \textbf{Method} & \textbf{mAP @ 0.5} \\
    \midrule
    Source only & 31.1 \\
    Domain Adaptive Faster R-CNN~\cite{chen2018domain} & 39.0 \\
    SPLAT-lite (ours) & \textbf{51.5} \\
    \midrule
    Oracle & 70.7 \\
    \bottomrule
    \end{tabular}
    \caption{
    We report mean average precision at a threshold of 0.5 IoU on the Cityscapes validation set.
    Our pixel-level adaptation model outperforms previous approaches, which limited themselves to feature-level adaptation, by a significant margin.
    In fact, we are able to recover over 50\% of the difference between the simple source-only baseline and the upper bound target oracle result.
    }
    \label{tab:detection}
\end{table}

\begin{figure}
    \centering
    \begin{subfigure}{\columnwidth}
    \centering
    \includegraphics[width=\columnwidth]{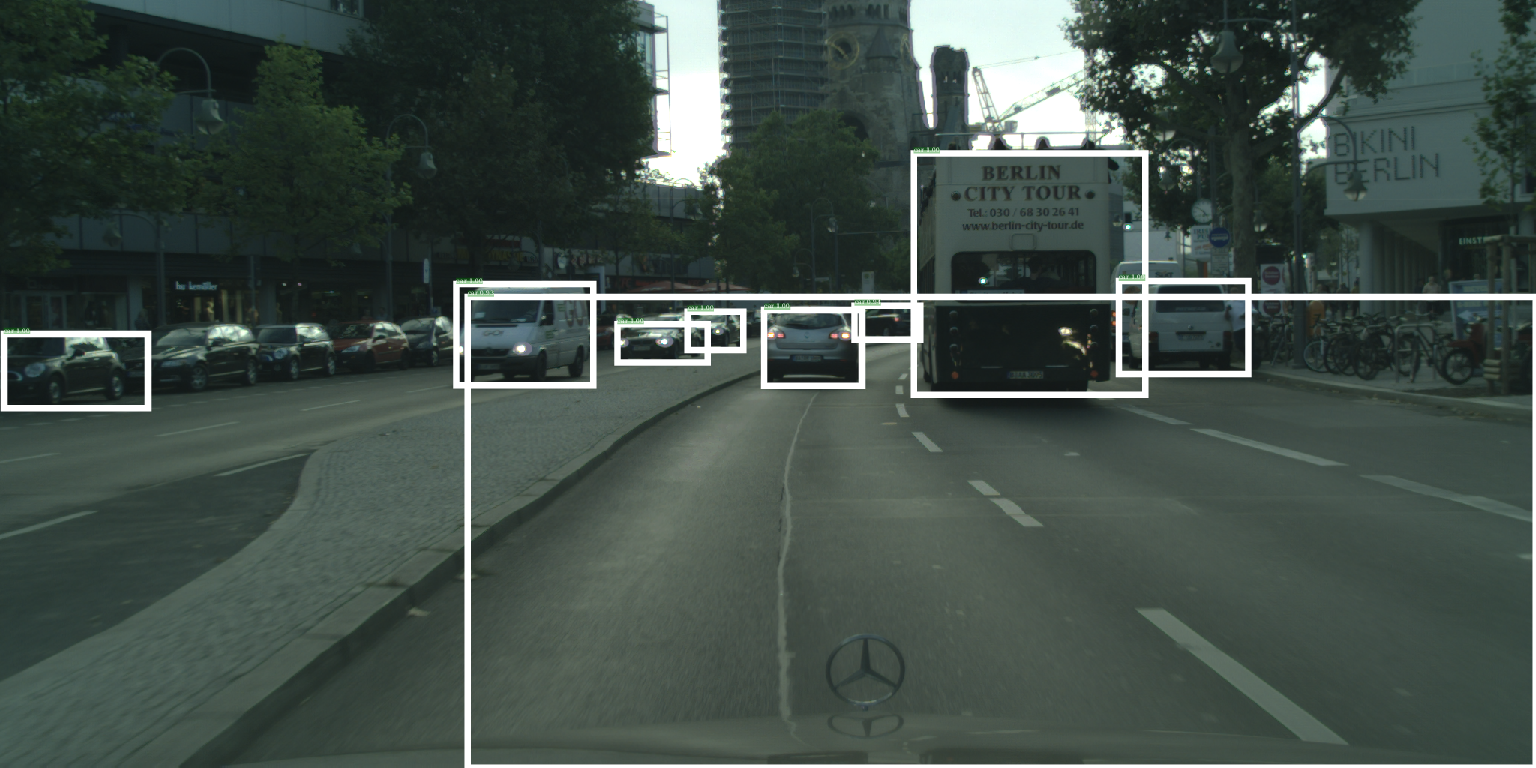}
    \caption{Before adaptation}
    \vspace{1em}
    \end{subfigure}
    \begin{subfigure}{\columnwidth}
    \centering
    \includegraphics[width=\columnwidth]{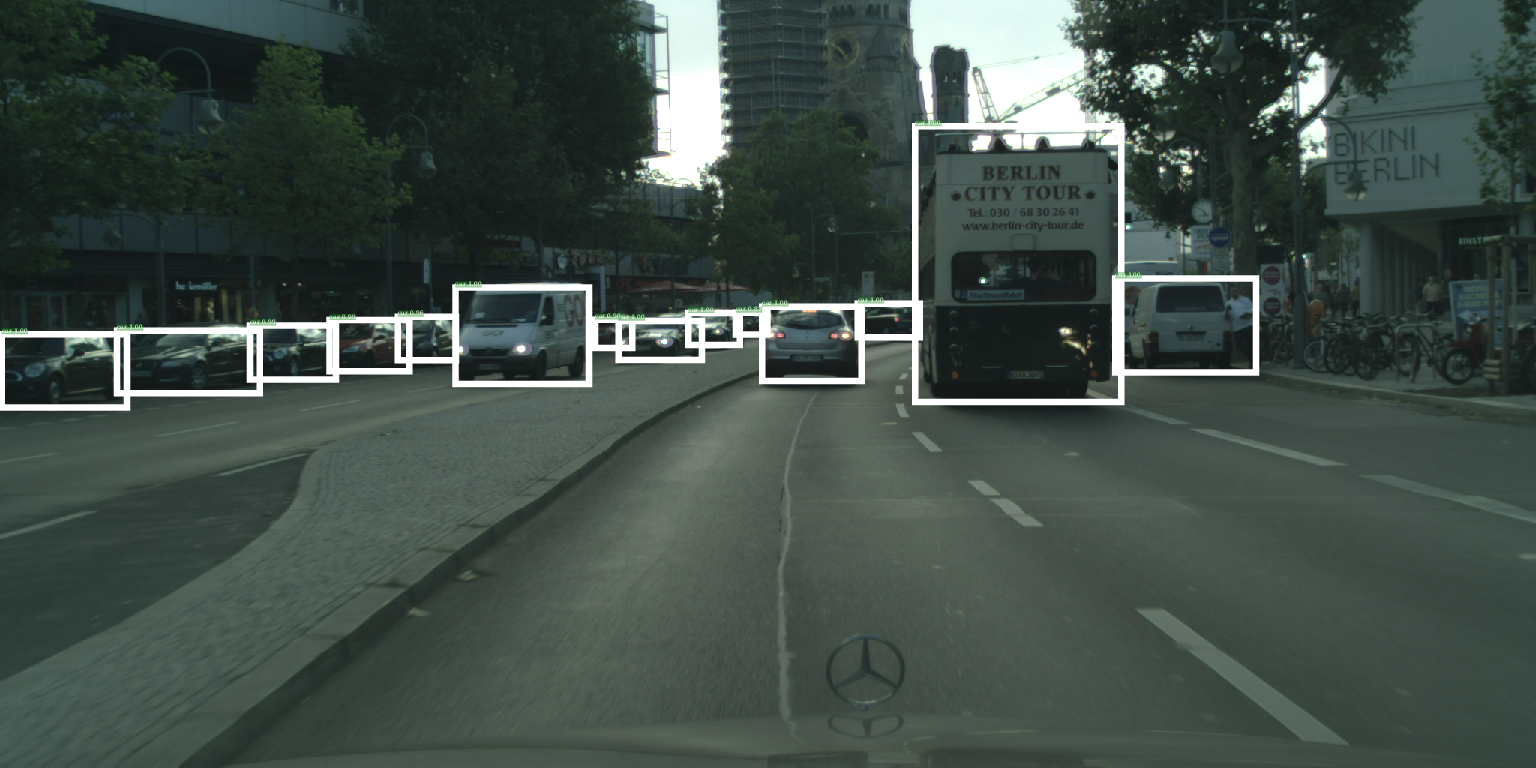}
    \caption{After adaptation}
    \end{subfigure}
    \caption{Comparing detections output by our model both before and after adaptation indicates that our model recognizes cars out of context and does not misattribute parts of cars, such as the hood ornament, for whole cars.}
    \label{fig:results-detection}
\end{figure}

We separately investigate the effects of the pseudo-label and aligned pair losses. Our results show that, for the detection adapation task, training on the pseudo-labeled dataset of source adapted images strictly outperforms training with a feature loss on aligned pairs. We first examine these losses independently, then explore combining both. All of our experiments are evaluated the validation set of \emph{Cityscapes}. 

When using these losses independently, we find that the pseudo-label loss outperforms the aligned pair loss by a significant margin: pseudo-labeling achieves 51.5 mAP versus the aligned pair loss, which achieves 43.3.
Somewhat surprisingly, combining the two losses appears to underperform the pseudo-label loss alone, achieving a mAP of only 47.5. We hypothesize that directly matching the features between source and target is too restrictive and prevents the model from properly learning features that work well in the target domain, and report final results using only the pseudo-label loss.

Our final results are presented in Table~\ref{tab:detection}.
Our proposed method achieves state-of-the-art performance, outperforming competing baselines by 12.5 points.
We also provide a comparison against a model trained using the fully-labeled Cityscapes training set, which we call the oracle model.
This serves as an upper bound for what level of performance we can reasonably expect to achieve via adaptation.
Comparing our SPLAT result against this oracle indicates that our model recovers over 50\% of the missing performance due to domain shift.

A qualitative analysis of detections both pre- and post-adaptation yields additional insight into our adaptation method.
We visualize detections produced by both the source-only baseline and our model on a sample image from the Cityscapes validation dataset in Figure~\ref{fig:results-detection}.
Before adaptation, the source-only model is decent at detecting cars that are driving in the center of the road.
However, it misses many of the cars that are parked along the side of the road, and it appears to be distracted by the presence of the car that the camera is mounted on along the bottom of the image.
Our adaptation method appears to fix both of these issues: almost every car parked on the side is detected, and no spurious detections are present.

\subsection{Pixel adaptation methods}
\label{sec:results-pixel}

\begin{table*}
    \centering
    \begin{tabular}{cccc}
    \toprule
    \textbf{Original} & \textbf{SPLAT-cycle} & \textbf{SPLAT-big} & \textbf{SPLAT-lite} \\
    \midrule
         \includegraphics[width=0.22\textwidth]{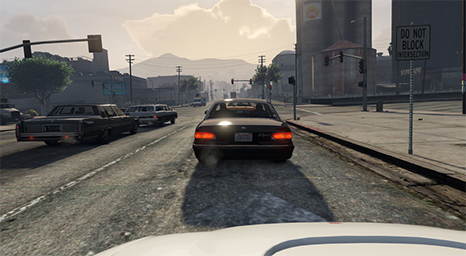} & 
         \includegraphics[width=0.22\textwidth]{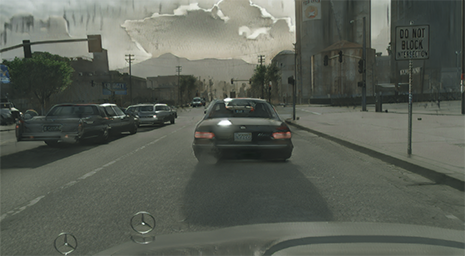} & 
         \includegraphics[width=0.22\textwidth]{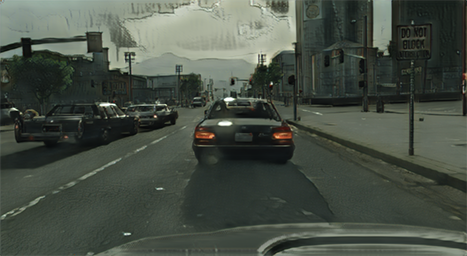} & 
         \includegraphics[width=0.22\textwidth]{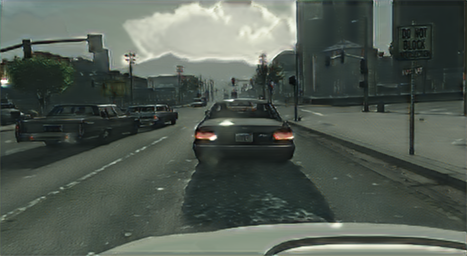} 
    \\
         \includegraphics[width=0.22\textwidth]{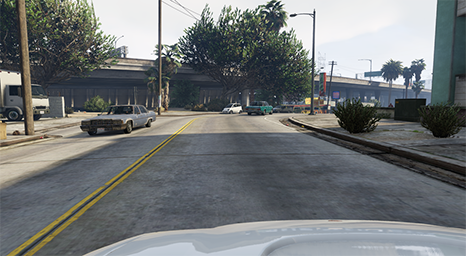} & 
         \includegraphics[width=0.22\textwidth]{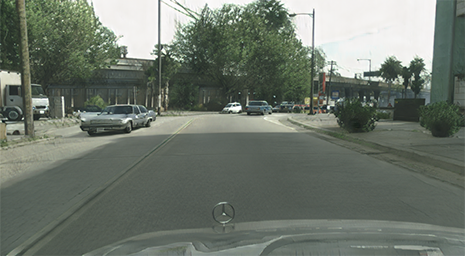} & 
         \includegraphics[width=0.22\textwidth]{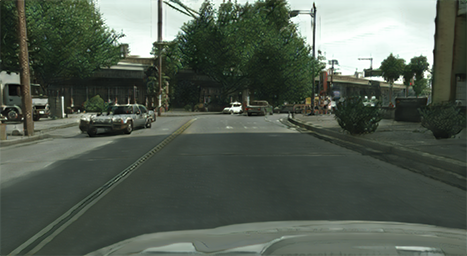} & 
         \includegraphics[width=0.22\textwidth]{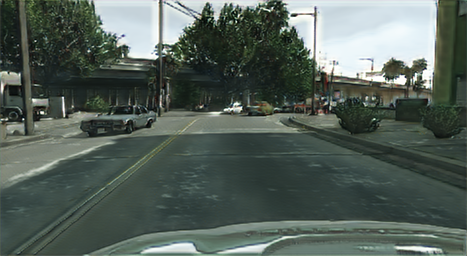} 
    \\
         \includegraphics[width=0.22\textwidth]{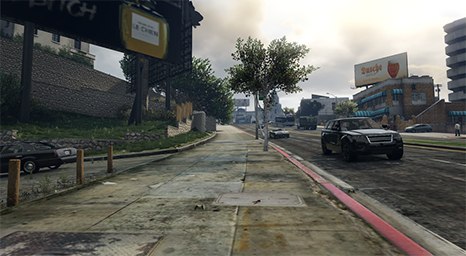} & 
         \includegraphics[width=0.22\textwidth]{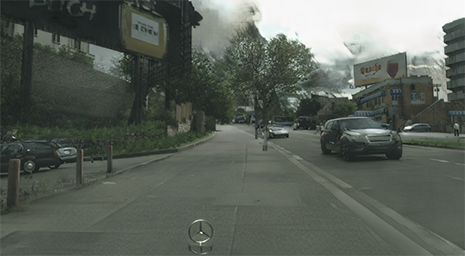} & 
         \includegraphics[width=0.22\textwidth]{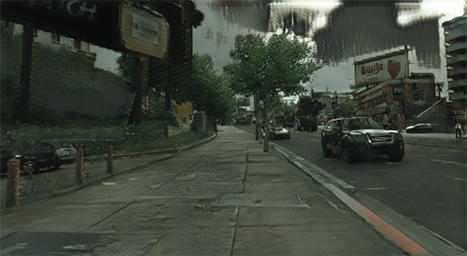} & 
         \includegraphics[width=0.22\textwidth]{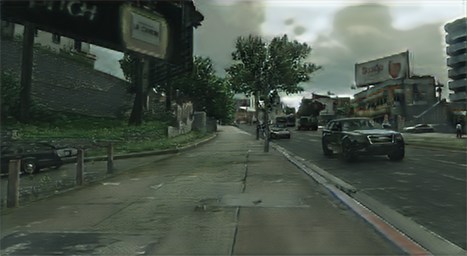} 
    \\
         \includegraphics[width=0.22\textwidth]{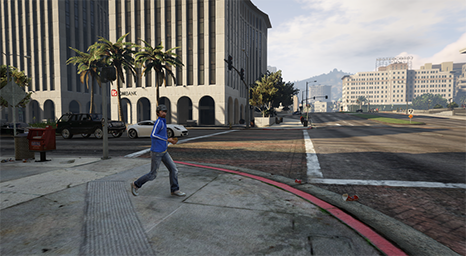} & 
         \includegraphics[width=0.22\textwidth]{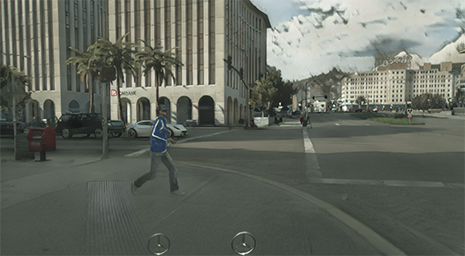} & 
         \includegraphics[width=0.22\textwidth]{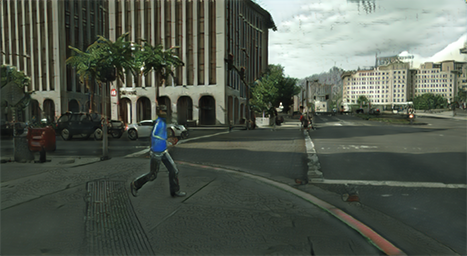} & 
         \includegraphics[width=0.22\textwidth]{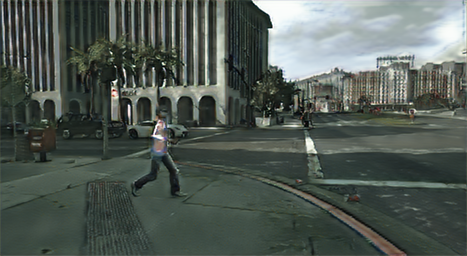} 
    \\
    \midrule
        \multicolumn{4}{c}{\textbf{Cityscapes}} \\
    \midrule
         \includegraphics[width=0.22\textwidth]{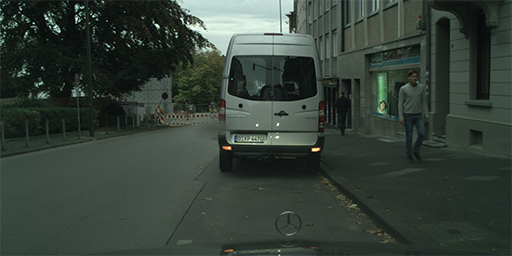} & 
         \includegraphics[width=0.22\textwidth]{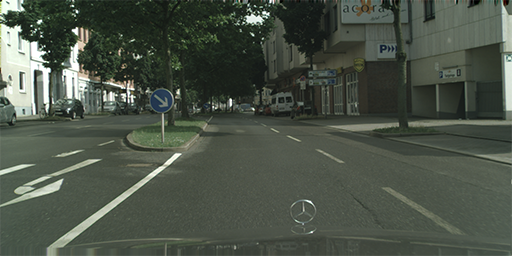} & 
         \includegraphics[width=0.22\textwidth]{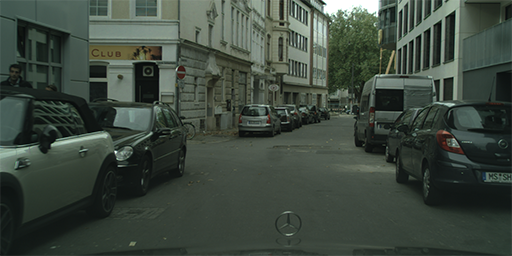} & 
         \includegraphics[width=0.22\textwidth]{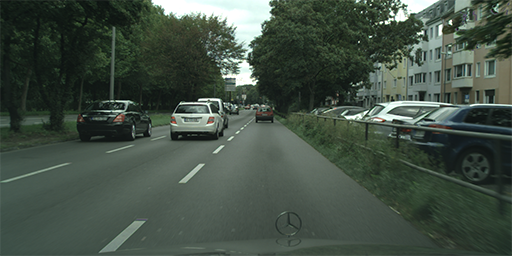} 
    \\
    \bottomrule
    \end{tabular}
    \caption{
    Side-by-side comparisons of pixel adaptation on source imagery.
    SPLAT-lite is able to generate effective target-style imagery while training almost 4 times faster than CyCADA.
    The target Cityscapes images are provided as examples of real target images for comparison, and do not directly correspond to the source images.
    }
    \label{tab:adapted-samples}
\end{table*}

The results presented in Section~\ref{sec:results-detection} were for a particular instantiation of SPLAT using our proposed cycle-free pixel adaptation method.
We refer to this instantiation of SPLAT as SPLAT-lite.
In this section, we explore the effect that our choice of pixel adaptation method has on our final detection performance.
We compare SPLAT-lite against a version of SPLAT trained using a CyCADA pixel transformer, which we refer to as SPLAT-cycle.
In doing so, we show that SPLAT is flexible enough to effectively use different pixel-adaptation methods while simultaneously validating our new cycle-free pixel transformation approach against these methods.
To demonstrate the efficiency of SPLAT-lite, we benchmark the time it takes to perform a training iteration on a single image (both forward and backward pass).
These timings use 256$\times$256 RGB images as input and were produced on a Tesla P100 GPU.

Additionally, we experiment with a larger, overparametrized variant of our cycle-free model that uses the same architectures as in Table~\ref{tab:splat-arch}, but with 8 times as many channels in each of its layers in both the generator and the discriminator.
We refer to this variant as SPLAT-big, and include it as a third version of SPLAT in our evaluation.

\begin{table*}
    \centering
    \small
    \begin{tabular}{lccc}
    \toprule
    \textbf{Method} & \textbf{Cityscapes mAP @ 0.5} & \textbf{Time per training iteration (s)} & \textbf{Speedup over SPLAT-cycle} \\
    \midrule
    SPLAT-cycle  & 48.1 & 0.377 & $\times$1.00 \\
    SPLAT-big    & 48.4 & 0.213 & $\times$1.77 \\
    SPLAT-lite   & \textbf{51.5} & \textbf{0.098} & $\times$\textbf{3.84} \\
    \bottomrule
    \end{tabular}
    \caption{
      We compare the effects of using different pixel adaptation methods on both final adaptation performance and speed during training.
      Training benchmarks are produced on a Tesla P100 with a 256$\times$256 image as input.
      Our proposed cycle-free pixel adaptation method SPLAT-lite has clear advantages over previous methods, both in the quality of the final model as well as the speed of the model during training.
    }
    \label{tab:benchmarks}
\end{table*}

Results from this full comparison are shown in Table~\ref{tab:benchmarks}.
We see that, in addition to achieving the strongest detection adaptation results in our framework, SPLAT-lite runs almost four times faster than SPLAT-cycle.
Despite how much more lightweight SPLAT-lite is, leveraging task information during training enables it to produce target-style images that are better suited for domain adaptation.

A comparison of SPLAT-lite against SPLAT-big is also illuminating.
Considering that SPLAT-lite is already able to produce effective adapted imagery, it is reasonable to suspect that SPLAT-big may be overparametrized.
In turn, an overparametrized generator is more likely to destroy the content of the original image during cross-domain image generation, since the increased number of parameters gives it more freedom to produce different outputs.
Nevertheless, we see that, although performance is worse than SPLAT-lite, the combination of the label alignment loss along with modern GAN stabilization techniques such as spectral normalization ensures that optimization proceeds in a stable manner, and the final result is competitive with cycle-based methods.

Examples of pixel-adapted source imagery are shown in Table~\ref{tab:adapted-samples}. By removing cycle-consistency constraints, SPLAT-big and SPLAT-lite are able to avoid learning domain artifacts that are irrelevant to the task at hand. In contrast, with cycle-consistency enforced, the pixel-adaptation model hallucinates the hood ornament that is common in the Cityscapes dataset. Notice, in column 2 and row 5, the hood ornament is even generated in unreasonable contexts, such as on the sidewalk. The final instantiation of our model, SPLAT-lite, also generates less noise than either of its highly-parameterized counterparts.

\section{Conclusion}

Our proposed method SPLAT is a novel approach to detector adaptation that utilizes pixel-level transforms to adapt from source to target domains.
SPLAT is a flexible model; it works on unlabeled target data, and when target labels are present, can additionally condition on them with a cycle-free loss to operate more accurately and efficiently.
Our model is also able to make use of unlabeled source data by inferring additional label information, thereby increasing the amount of training data available to learned task networks.
By incorporating these improvements, our final model adapts images 3.8 times faster than previous cycle-based methods while improving 12.5 mAP over previous methods on a challenging detection adaptation setting.

{\small
\bibliographystyle{ieee}
\bibliography{references}
}

\end{document}